\documentclass[10pt,twocolumn,letterpaper]{article}

\usepackage[pagenumbers]{cvpr} % To force page numbers, e.g. for an arXiv version

\usepackage{bm}

\usepackage{amsmath,amssymb,amsfonts}
\usepackage{graphicx}
\usepackage{booktabs}
\usepackage{algorithm}
\usepackage{algpseudocode}
\usepackage{subcaption}        % OK; avoid 'subfigure'
\usepackage{placeins}
\usepackage{makecell} 

\definecolor{cvprblue}{rgb}{0.21,0.49,0.74}
\usepackage[pagebackref,breaklinks,colorlinks,allcolors=cvprblue]{hyperref}

\def\paperID{17547} % *** Enter the Paper ID here
\def\confName{CVPR}
\def\confYear{2026}

\title{Multi-Sensor Matching with HyperNetworks}

\author{Eli Passov\\
Faculty of Computer Science\\
Bar-Ilan University\\
Ramat-Gan, Israel\\
{\tt\small elipassov@gmail.com}
\and
Nathan S. Netanyahu\\
Faculty of Computer Science\\
Bar-Ilan University\\
Ramat-Gan, Israel\\
\and
Yosi Keller\\
Faculty of Engineering\\
Bar-Ilan University\\
Ramat-Gan, Israel\\
}

\begin{document}
\maketitle
\begin{abstract}
Hypernetworks are models that generate or modulate the weights of another
network. They provide a flexible mechanism for injecting context and task
conditioning and have proven broadly useful across diverse applications
without significant increases in model size. We leverage hypernetworks to
improve multimodal patch matching by introducing a lightweight
descriptor-learning architecture that augments a Siamese CNN with (i)
hypernetwork modules that compute adaptive, per-channel scaling and shifting
and (ii) conditional instance normalization that provides modality-specific
adaptation (e.g., visible vs.\ infrared, VIS–IR) in shallow layers. This combination preserves
the efficiency of descriptor-based methods during inference while increasing
robustness to appearance shifts. Trained with a triplet loss and
hard-negative mining, our approach achieves state-of-the-art results on
VIS-NIR and other VIS-IR benchmarks and matches or surpasses prior methods on
additional datasets, despite their higher inference cost. To spur progress on
domain shift, we also release GAP-VIR, a cross-platform (ground/aerial) VIS-IR
patch dataset with 500K pairs, enabling rigorous evaluation of cross-domain
generalization and adaptation. We make our code publicly available at \footnote{
\url{https://anonymous.4open.science/r/multisensor_hypnet-6EE1}.}

\end{abstract}

\section{Introduction}

\label{sec:introduction}

\begin{figure}[t]
\vspace{-0mm}
\centering\includegraphics[width=\columnwidth]{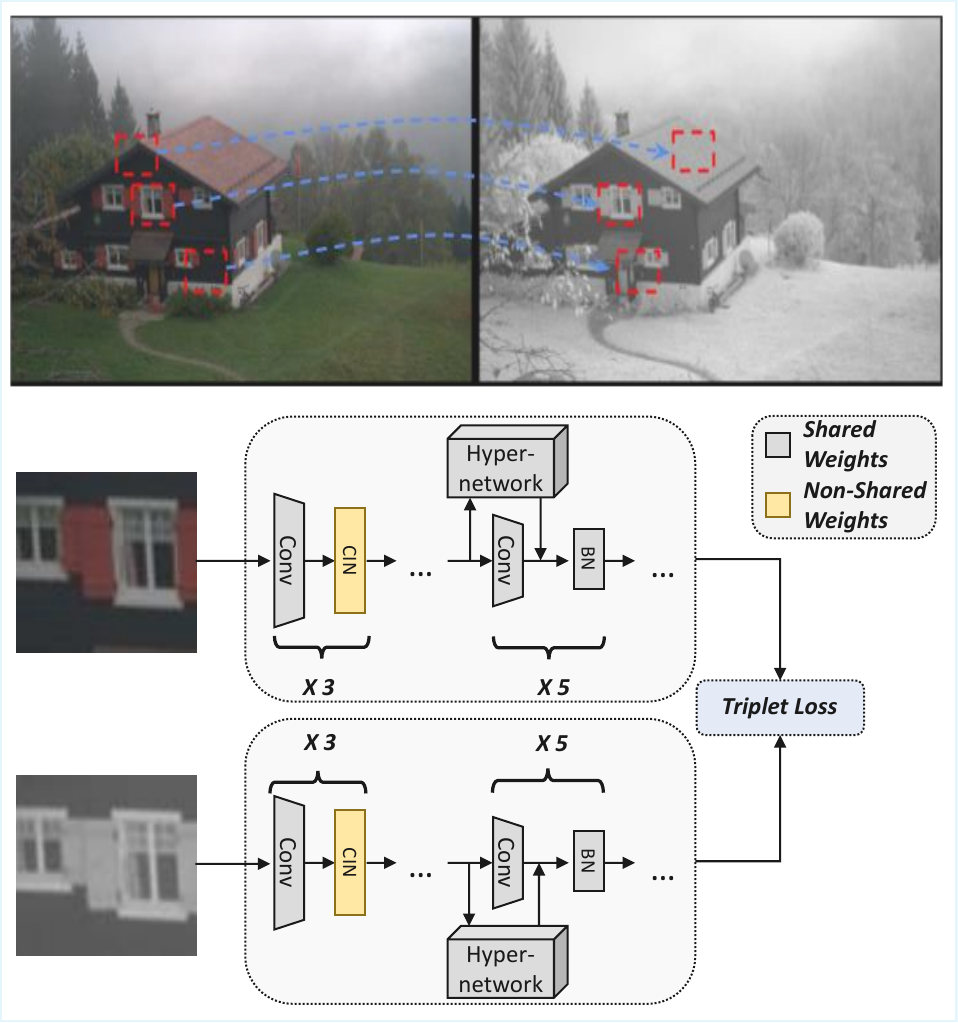}
\vspace{-0mm}\caption{Top: Patch matching between visible (grayscale) and NIR images. Bottom: The proposed hypernetwork-based Siamese CNN, where each branch encodes patches into descriptors. Most weights are shared; deeper layers include convolutions, hypernetwork modules, and batch normalization (BN), while shallow layers use convolutions and conditional instance normalization (CIN) for modality-specific adaptation. A symmetric triplet loss trains the network to learn discriminative descriptors.}%
\label{fig:network_diagram}%
\vspace{-0mm}
\end{figure}

Patch matching is a fundamental task in computer vision with numerous
applications, including image registration
\cite{wang2018deep, ye2022robust}, image retrieval \cite{philbin2007object,
dou2022reliability}, and simultaneous localization and mapping (SLAM)
\cite{shi2018planematch, guo2023descriptor}. The patch matching
task involves determining whether two image patches originate from the same
physical area of interest. In multimodal patch matching, patches come from
different modalities, such as RGB and near-infrared (NIR). Multimodal
patch matching is used in medical imaging
\cite{sotiras2013deformable} and multisensor image alignment
\cite{irani1998robust}. Modal differences are challenging because the relationship between pixel intensities and the appearance of objects and boundaries varies across modalities. However, these differences make using a
network pretrained on large RGB datasets (e.g., ImageNet
\cite{deng2009imagenet}) challenging—if not impractical—for multimodal patch matching.

Earlier works \cite{lowe2004distinctive, chen2009real, firmenichy2011multispectral, aguilera2012multispectral, aguilera2015lghd} employed classical feature extraction methods, such as the Scale-Invariant Feature Transform (SIFT), Difference of Gaussians (DoG), or Gabor filters, to create meaningful patch features known as descriptors.
Applying metrics such as Euclidean distance to pairs of descriptors enabled determining whether two
patches match. Recent methods replace classical feature extractors with deep
convolutional neural networks (CNNs) \cite{balntas2016pn,
aguilera2017cross, en2018ts, baruch2021joint, wang2019better} to improve the
quality of descriptors. Many works use Siamese CNNs \cite{jahrer2008learned},
sharing the architecture and weights between the two modalities. Others use pseudo-Siamese \cite{baruch2021joint} CNNs, where the architecture is shared but all or some weights are learned independently for each
modality. Descriptor networks are trained using various losses based on
Euclidean distance, such as hinge loss \cite{simo2015discriminative} or
triplet loss \cite{hoffer2015deep, balntas2016learning, wang2019better}. Some
works \cite{wang2019better,tian2017l2,mishchuk2017working} utilize hard
negative mining, which continually selects the hardest non-matching patches
for training. Besides descriptor learning, some methods employ
similarity learning (often called metric
learning)~\cite{han2015matchnet,zagoruyko2015learning,aguilera2016learning}.
In these works, patches from both modalities are typically processed jointly
by a single CNN to determine if they match. Furthermore, some
works~\cite{quan2019afd,ko2023spectral,yu2022multibranch,yu2023feature,yu2025relational}
propose architectures with additional layers and branches that connect to and
draw from intermediate feature extraction layers. Although similarity-based
methods have achieved favorable results in recent years, using them for
inference is less practical \cite{jiang2021review,moreshet2021paying} due to
their higher computational complexity, as discussed in
Section~\ref{subsec: descriptor vs metric}.

In this work, we present a new \textit{descriptor learning} approach to
multimodal patch matching, which enhances the Siamese network for feature
extraction using two key components. First, we introduce a hypernetwork
\cite{ha2016hypernetworks} module that captures contextual information for
each channel. An encoder–decoder network then computes dynamic scaling and shifting from this context. Second, we introduce
modality-specific context via Conditional
Instance Normalization (CIN) \cite{dumoulin2016learned}, originally used for
style transfer. We show that applying CIN modules in the shallow layers of a feature extraction network allows the model to more flexibly adapt to different modalities. The introduction of modality-based components converts a fully Siamese network into a pseudo-Siamese network, where nearly all weights, except for a small fraction, are shared between the two modalities. We demonstrate how these additions enhance the performance of widely used descriptor learning architectures for multimodal patch matching. We apply both modules as illustrated in Fig.~\ref{fig:network_diagram}.
In the shallow layers, CIN is applied to enable modality-specific
adaptation. Hypernetwork blocks are introduced in the deeper layers, where they
enrich the shared feature context. We train the Siamese network with a
triplet loss~\cite{hoffer2015deep} combined with hard negative
mining~\cite{mishchuk2017working}. This setup achieves state-of-the-art (SOTA) results on the
VIS-NIR~\cite{aguilera2016learning} dataset as well as other common benchmarks. In
addition, we conduct ablation studies to analyze the contribution of each
proposed module and to compare alternative backbones, including
ResNet-18~\cite{he2016deep}, ViT~\cite{dosovitskiy2020image}, and
Swin~\cite{liu2021swin}.

Finally, given our results on previously used datasets, we observe that
recent work has reached saturation. The lack of new multimodal patch datasets hinders
algorithm development and the study of key challenges, notably domain
transfer and domain adaptation. To address this, we introduce the
\textbf{G}round \textbf{A}erial \textbf{P}atch \textbf{V}is-\textbf{IR}
(GAP-VIR) dataset (details in Appendix~A), containing
500K pairs of matched and unmatched VIS-IR patches from cross-platform imagery
(ground-based and aerial). GAP-VIR uniquely enables testing of cross-modal and
cross-platform training, as well as evaluating cross-platform adaptation. Our research offers the following key contributions:

\begin{itemize}
\item A novel lightweight hypernetwork module
infuses global context into feature extraction networks, thereby
improving descriptor learning methods.

\item The proposed CIN modules improve multimodal matching accuracy by
infusing modality-specific context into the network.

\item We achieve state-of-the-art results on multimodal patch matching with
the proposed approach.

\item GAP-VIR, a new cross-platform multimodal (VIS-IR) patch-matching
dataset, is introduced to facilitate future research.
\end{itemize}

\section{Related Work}

\label{sec:Related Work}

Before CNN-based approaches, patch matching primarily relied on handcrafted, appearance-invariant features. Keller et al.~\cite{keller2002robust} aligned gradient maxima for affine registration through iterative optimization. A similar approach~\cite{keller2006multisensor} detected false correspondences and adaptively reweighted them using directional similarity. Irani et al.
\cite{irani1998robust} used directional derivative magnitudes as robust descriptors and optimized matching through a coarse-to-fine correlation strategy. Netanyahu et al.~\cite{netanyahu2004georegistration} employed iterative matching of multiresolution wavelet features for large-scale image registration, using the partial Hausdorff distance for robust correspondence. Lemoigne et al.~\cite{lemoigne2011image} developed a modular framework for evaluating multitemporal and multisensor registration, integrating diverse features, similarity measures, and optimization strategies.
Other works proposed modified SIFT descriptors to enhance geometric invariance and local discriminability. Chen et al.~\cite{chen2009real} introduced a symmetric SIFT descriptor that is scale and rotation invariant. Ma et al.~\cite{ma2010mi} developed a mirror-invariant variant, and Hasan et al.~\cite{hasan2012modified} improved spatial resolution through gradient thresholding and sub-windowing.  Kupfer et al.~\cite{kupfer2014efficient} proposed an efficient SIFT-based algorithm for subpixel registration that filters outliers by mode seeking over keypoint scale ratios, rotation differences, and vertical shifts.
Beyond SIFT-based techniques, Aguilera et al.~\cite{aguilera2012multispectral} used DoG-based keypoints and edge-oriented histograms that capture shape and contour information, later refined with log-Gabor filters~\cite{aguilera2015lghd}. This approach was refined in \cite{aguilera2015lghd} by using log-Gabor filters to compute the descriptors. Zhu et al. \cite{zhu2023r2fd2} used log-Gabor filters with multichannel auto-correlation to detect interest points and construct a rotation-invariant descriptor. The
self-similarity descriptor by Shechtman et al. \cite{shechtman2007matching}
is computed per patch, image, or video, and corresponding pairs are compared to identify matches. Kim et al.~\cite{kim2015dasc} extended this idea to a dense adaptive self-correlation descriptor that measures similarity between randomly sampled patches within a local support window. Ye et al.~\cite{ye2022robust} further addressed nonlinear radiometric and geometric distortions using steerable-filter descriptors based on first- and second-order gradients.

Siamese networks extract feature representations of two
inputs in parallel through shared weights and are widely used in descriptor learning.
Simo-Serra et al. \cite{simo2015discriminative} utilized the same CNN for both modalities, coupled with a hinge loss that maximizes the distance between non-matching pairs. Other works
\cite{balntas2016pn,wang2019better,ma2021image,tian2020hynet,balntas2016learning}
use a triplet loss \cite{hoffer2015deep}, which also minimizes the distance of
matching pairs. Aguilera et al. \cite{aguilera2017cross} introduced the
quadruplet loss, where all distances between the features of two sets
of matching pairs are included in a loss function that minimizes or maximizes them depending on whether the pairs match or not. Ben-Baruch et al. \cite{baruch2021joint} used a hybrid approach of Siamese and pseudo-Siamese networks, sharing only some of their weights. Other works
\cite{wang2019better,tian2017l2,mishchuk2017working} used hard negative mining,
where negative samples are selected for hinge or triplet losses. Zhou et al.
\cite{zhou2023deep} combined hard negative mining with continual learning to obtain rotation-invariant features. Kumar et al.
\cite{kumar2016learning} proposed a global loss that minimizes the mean distance of matching features, maximizes the mean distance of non-matching features, and minimizes the variance within both groups.

Moreshet et al. \cite{moreshet2021paying} added a spatial pyramid pooling
layer on top of the backbone and a transformer layer with learned positional
embeddings to better capture spatial feature relationships. Tian et al.
\cite{tian2019sosnet} introduced a second-order similarity regularization to enforce similarity between distances of different patch pairs. The same authors \cite{tian2020hynet} conducted an
analysis of the gradient with $L_{2}$ and dot-product distances. They applied
a regularization term to the positive descriptor distances to tackle the
impact of image intensity variability on the descriptors. The authors used a
hybrid dot-product and $L_{2}$ similarity measure in the main loss function.
They also employed a filter response normalization (FRN) layer followed by a
threshold linear unit (TLU), both introduced in \cite{singh2020filter}. Han
et al. \cite{han2015matchnet} proposed a convolutional Siamese network that
extracts features from two patches, concatenates them, and passes the result into a
classification MLP (Multi-Layer Perceptron). Zagoruyko
\cite{zagoruyko2015learning} and Aguilera \cite{aguilera2016learning}
considered three architectures: Siamese, pseudo-Siamese, and two-channel
networks. In the hybrid approach of Quan et al. \cite{quan2019afd}, the two
inputs are passed through separate branches of a Siamese CNN. The same authors
\cite{quan2021multi} also proposed extracting the concatenation, subtraction,
and multiplication of the output features of each modality for every layer.
Each of the three relational features is passed through a small
channel attention network and trained using an LMCL loss. Ko et al.
\cite{ko2023spectral} proposed a similarity-learning approach in which two
modality-specific networks first convert inputs into the opposite modality.
Then, for each modality, the original and converted inputs are passed into a
Siamese network to produce a pair of feature vectors, which are used to
compute a similarity score. In a series of works, Yu et al.~\cite{yu2022multibranch,yu2023efficient,yu2025relational} first applied channel attention blocks for feature extraction, then used an encoder–decoder architecture to regularize and enhance the extracted features, and finally introduced a cross-spectral attention network on top of the Siamese feature-extraction network.

Multimodal frameworks such as MINIMA~\cite{ren2024minima} and
MatchAnything~\cite{he2025matchanything} shift focus from per-pair similarity
networks toward large-scale modality-invariant pretraining. These approaches
achieve cross-modal generalization by leveraging synthetic or diverse
multimodal datasets. Yu et al.~\cite{yu2024and} proposed a hybrid descriptor-
and similarity-learning approach trained with hard-negatively mined samples.
Zhang et al.~\cite{zhang2023ssml, zhang2023learning} used a quadruplet
network, passing patches through two Siamese branches and two pseudo-Siamese
branches before concatenating the outputs. To further enhance feature
representation, they incorporated Squeeze-and-Excitation (SE)
\cite{hu2018squeeze}, along with either a transformer or Coordinate Attention
(CA) \cite{hou2021coordinate} for channel attention, in processing intermediate
feature maps. Transformer-based cross-modal matchers such as
XoFTR~\cite{tuzcuouglu2024xoftr} and modality-unifying networks for VIS–IR
re-identification~\cite{yu2023modality} highlight the role of attention in
bridging spectral appearance gaps. Our work complements these approaches
by focusing on lightweight architectural improvements for VIS-IR descriptor learning.

On the dataset side, there are additional multisensor datasets such as
msGFM~\cite{han2024bridging}, a cross-modal ship re-identification
benchmark~\cite{wang2025cross}, and large-scale RGB-D datasets for
reconstruction~\cite{voynov2023multi}. In addition, Ran et
al.~\cite{ran2025diffv2ir} introduced IR500K, a large-scale aggregation of
VIS–IR pairs from multiple sources that serves as the foundation for our GAP-VIR dataset.

\subsection{Hypernetworks}

\label{subsec:hypernetworks}

Hypernetworks (introduced in \cite{ha2016hypernetworks}) are used across a wide range of
problems, and their architectures and implementation details vary widely depending on the application. For example, \cite{volk2023example} uses a hypernetwork
to adjust classifier weights based on an embedding produced
by another network for multi-source adaptation to unseen domains.
In \cite{lutati2023ocd}, a hypernetwork conditioned on a single input of the
target network adapts its weights to improve performance in image classification and 3D reconstruction tasks. Other applications
include semantic segmentation \cite{nirkin2021hyperseg}, implicit neural
representations \cite{chen2022transformers, sitzmann2020implicit}, 3D shape
reconstruction \cite{littwin2019deep}, and continual learning
\cite{von2019continual} across diverse domains. A comprehensive review and taxonomy of hypernetworks are provided in \cite{chauhan2023brief}.

\subsection{Descriptor Learning vs. Similarity-Based Methods for Patch
Matching}

\label{subsec: descriptor vs metric}

Patch matching methods can be broadly categorized into descriptor-learning and
similarity-based approaches, which differ substantially in computational cost.
When matching patches between two groups of $n$ samples, descriptor learning
requires running network inference $n$ times per group to extract
feature vectors. After extraction, the pairwise distances between all
descriptors from both groups are computed to perform matching, with a
worst-case complexity of $O(n^{2})$, which can be reduced to $O(n \log n)$
using efficient nearest-neighbor search algorithms, such as kd-trees.
In contrast, similarity-based methods require running neural network
inference for all $O(n^{2})$ matching scores. As a result, descriptor learning
is generally more scalable and practical for large datasets and real-time
applications.

\begin{figure}[tb]
\centering\includegraphics[width=\columnwidth]{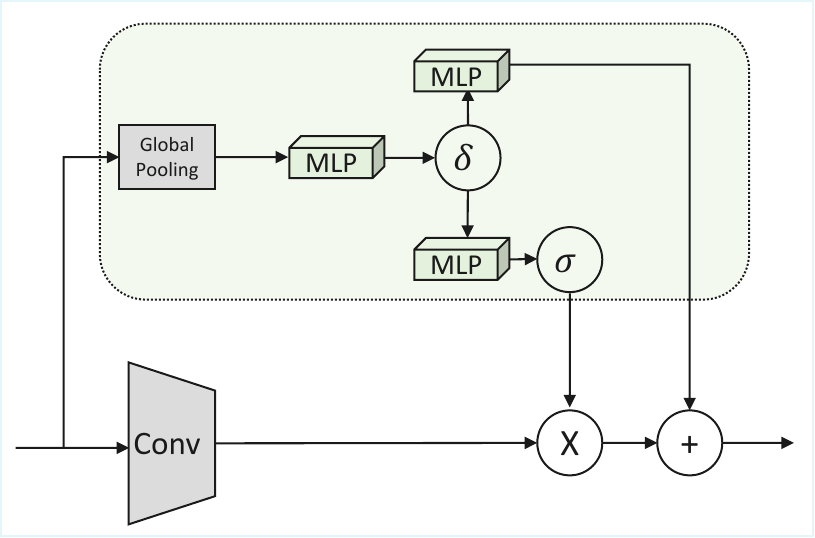}
\caption{\textbf{Hypernetwork module.}
The architecture of the hypernetwork: the convolution input is passed through
global pooling, followed by fully connected nonlinear layers that produce the
scaling and shifting factors. These factors are applied to the convolution
output after being replicated across spatial dimensions.}
\label{fig:hypernet_architecture}%
\vspace{-0mm}
\end{figure}

\section{Patch Matching using Hyper-Networks}

\label{sec:Proposed Method}

We propose a patch-matching scheme based on a hypernetwork architecture that
improves matching accuracy and robustness to appearance variations. Our
solution is a descriptor-learning method built on a convolutional Siamese
neural network. Given a pair of visual and infrared patches, $\mathbf{x_{1}}$ and $\mathbf{x_{2}}$, respectively, the network produces corresponding descriptor
vectors, $\mathbf{d_{1}}$ and $\mathbf{d_{2}}$, respectively, as illustrated in
Fig.~\ref{fig:network_diagram}. Patch matching is determined by the distance
between the descriptor vectors.

The Siamese backbone, described in Section~\ref{subsec_network_architecture},
is convolutional and shares most weights between branches, except for a few components. The
hypernetwork modules, detailed in Section~\ref{hypernetwork_block}, are added
to the deeper layers of the network to enable image-adaptive parameter learning
that responds dynamically to appearance variations between the visual and IR images. We
train the network using a triplet margin loss, detailed in
Section~\ref{ref:loss_function}.

\subsection{Network Architecture}
\label{subsec_network_architecture}

The proposed Siamese network is shown in Fig.~\ref{fig:network_diagram} and
Table~\ref{tbl:network_architectures}. It consists of eight convolutional
layers, chosen to ensure a fair comparison with previous SOTA methods~\cite{moreshet2021paying}.

Batch normalization (BN) is used in deeper layers, while conditional instance normalization (CIN) is applied in the first three layers. The CIN blocks are not weight-shared between the Siamese branches. Breaking weight symmetry allows modality-specific adaptation to the distinct characteristics of visible and IR patches.

The hypernetwork modules are integrated on top of the last five convolutional
layers. This design enhances the network’s ability to handle appearance
variations between objects in visual and NIR patches. Within the Siamese
network, all weights are shared between the visible and NIR branches, except
for the CIN block parameters. The effectiveness of the CIN and
hypernetwork modules is validated through an ablation study
(Section~\ref{subsec:ablation_studies}).

\subsection{Hypernetwork Module}

\label{hypernetwork_block}

\begin{table}%[tb]

\caption{Hypernetwork module architecture.}%
\vspace{-0mm}
\label{tbl:hypernetwork_block_architecture}
\begin{center}
{\small
\begin{tabular}{lcc}
\toprule
\textbf{Part} & \textbf{Module} & \textbf{Output size} \\
\midrule
{Input} &  & $C_{in} \mathbin{\! \times \!} H \mathbin{\! \times \!} W$ \\
\midrule
{Bottleneck} & Global pooling & $C_{in}$ \\
 & FC & $\tfrac{C_{in}}{8}$ \\
\midrule
Scaling & FC + Sigmoid & $C_{out}$ \\
\midrule
Shifting & FC & $C_{out}$ \\
\bottomrule
\end{tabular}
}
\end{center}
\vspace{-0mm}
\end{table}

The hypernetwork module $H$, illustrated in
Fig.~\ref{fig:hypernet_architecture}, computes the per-channel scaling and
shifting factors applied to the convolutional layer output $L(\mathbf{x})$.
Given $\mathbf{x} \in \mathbb{R}^{C_{in} \times H_{in} \times W_{in}}$ as
input, $L$ produces the output feature map
$\mathbf{y}\in\mathbb{R}^{C_{out} \times H_{out} \times W_{out}}$. The input $\mathbf{x}$ to $L$ is also provided to the hypernetwork module.
%Note that $H_{out}, W_{out}$ may differ from $H_{in}, W_{in}$ if the convolution involves downsampling.

% \begin{figure}[ptb]
% \centering\includegraphics[width=\columnwidth]{figures/network_architecture.pdf}\caption{\textbf{Siamese CNN architecture.} The network shares most weights across branches, with deeper layers containing convolutions, hypernetwork modules, and batch normalization (BN). The shallower layers include convolutions and conditional instance normalization (CIN), the only component not shared between branches. A symmetric triplet loss is used during training to learn discriminative descriptors.}

% \label{fig:full_architecture}%
% \end{figure}

\begin{table}[tb]
\caption{CNN architecture used in Hyp-Net: a convolutional backbone followed
by convolutional layers with hypernetwork modules, ending with a fully
connected head. Input size: $64\times64$ patch. Conv: $3\times3$
convolution; Hyp(Conv): hypernetwork module with a $3\times3$ convolution. Each
layer ends with a GELU nonlinearity~\cite{hendrycks2016gaussian}.}%
\label{tbl:network_architectures}
\centering
\small
\begin{tabular}{lccc}
\toprule
\textbf{Layer} & \textbf{Channels} & \textbf{Stride} & \textbf{Dilation} \\
\midrule
Conv+CIN+GELU  & 32 & 1 & 1 \\
Conv+CIN+GELU  & 32 & 2 & 1 \\
Conv+CIN+GELU  & 64 & 1 & 2 \\
\midrule
Hyp(Conv)+BN+GELU  & 64 & 2 & 1\\
Hyp(Conv)+BN+GELU  & 128 & 1 & 2 \\
Hyp(Conv)+BN+GELU  & 128 & 2 & 1\\
Hyp(Conv)+BN+GELU  & 128  & 1 & 1 \\
Hyp(Conv)+GELU  & 128 & 1 & 1 \\
\midrule
Flatten & 8192 & & \\
Dropout (0.5) & 8192 & & \\
FC & 128 & & \\
\bottomrule
\end{tabular}
\vspace{-0mm}
\end{table}

\subsection{Loss Function}
\label{ref:loss_function}

The network is trained using the triplet margin loss~\cite{hoffer2015deep}:
\begin{equation}
\mathcal{L}(d_{a}, d_{p}, d_{n}) = \max\left( 0,\, D(d_{a}, d_{p}) -
D(d_{a}, d_{n}) + \alpha \right),
\end{equation}
where $d_{a}$ is the anchor sample, $d_{p}$ and $d_{n}$ are the positive and
negative samples, respectively, $\alpha$ is the margin parameter (set to 1
in our experiments), and $D$ denotes the distance metric. Specifically, we use the
squared $\mathcal{L}_{2}$ norm:
\begin{equation}
D(d_{1}, d_{2}) = \left\| \frac{d_{1}}{\| d_{1} \|} -
\frac{d_{2}}{\| d_{2} \|} \right\|^{2}.
\end{equation}

Given a batch of $N$ matching patch pairs from both modalities, $(\mathbf{x}_{1}^{1}, \mathbf{x}_{1}^{2}), \dots, (\mathbf{x}_{N}^{1}, \mathbf{x}_{N}^{2})$, the network produces the corresponding descriptor pairs ${(d_{1}^{1}, d_{1}^{2}), \dots, (d_{N}^{1}, d_{N}^{2})}$. For a given anchor, the positive sample is its paired descriptor, and the negative sample is a non-matching descriptor from the other modality.
%, i.e., $d_{n(i,k)} = d_j^l \mid \min\limits_{j \in \left\{1 \dots N\right\} \setminus \{i\}} D\left(d_i^k, d_j^l\right)$.

% \vspace{-3mm}

\section{GAP-VIR Dataset}

\label{gap_vir_dataset}

\begin{figure}[tb]
\centering
\begin{minipage}{0.48\textwidth}
\centering
\includegraphics[width=\linewidth]{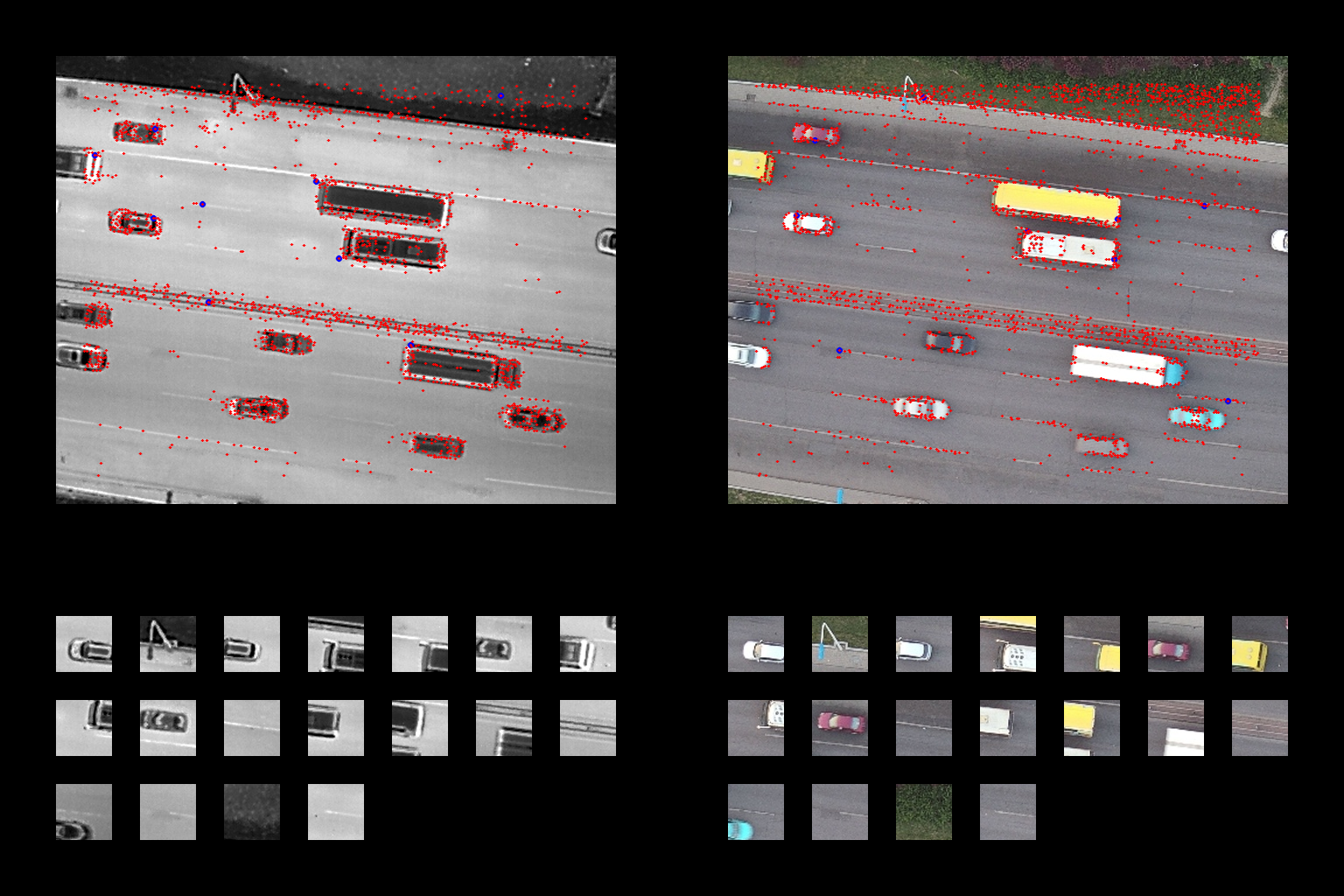}\\
\textbf{(a)} Aerial footage
\end{minipage}\hfill\begin{minipage}{0.48\textwidth}
\centering
\includegraphics[width=\linewidth]{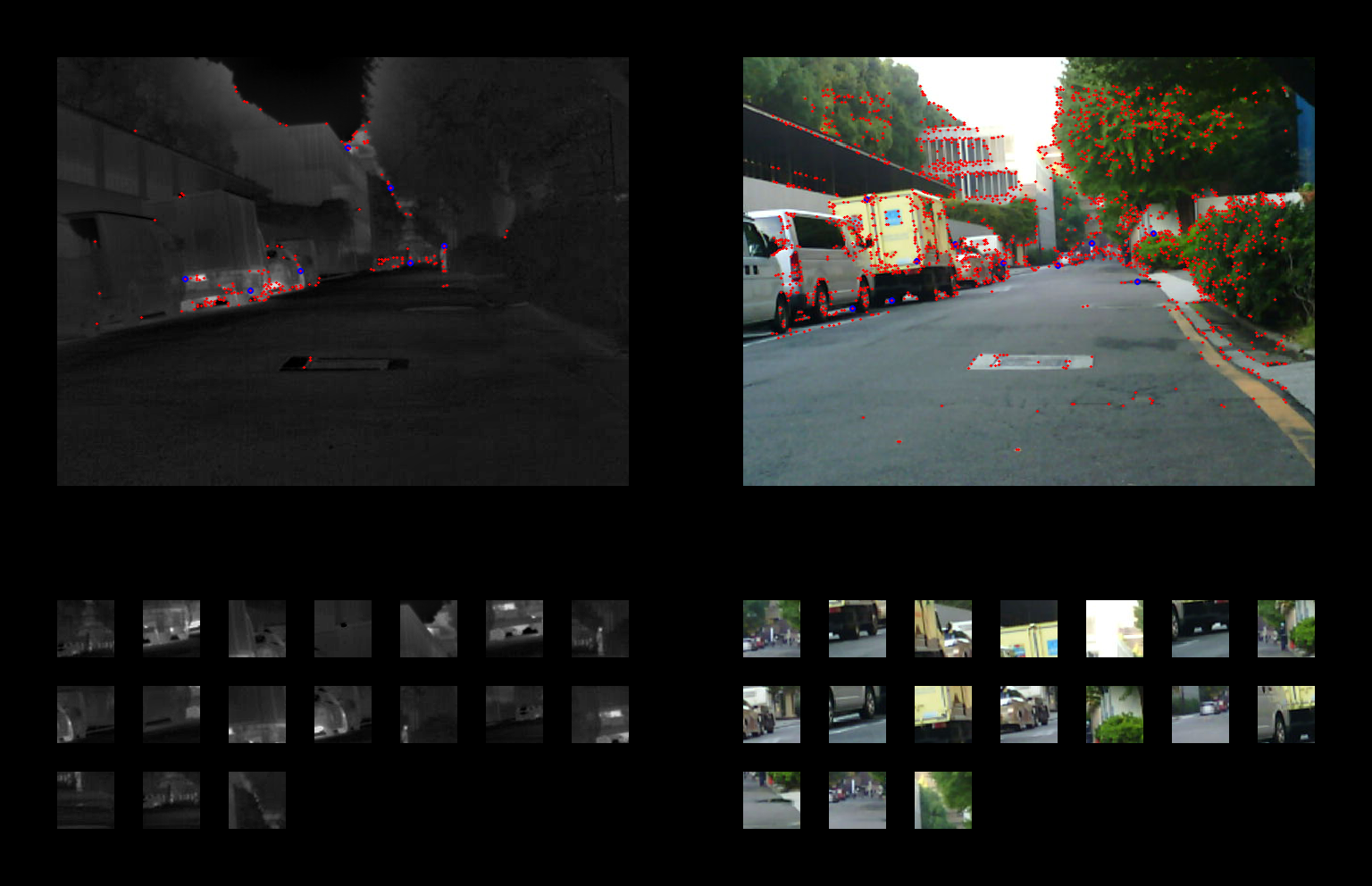}\\
\textbf{(b)} Ground footage
\end{minipage}
\caption{Patch extraction from paired aerial (a) and ground (b) imagery. Top:
FAST keypoints (red) and selected keypoints (blue) overlaid on IR (left) and
RGB (right) images. Bottom: corresponding $64\times64$ patches from both
modalities. The detailed patch extraction procedure is provided in
Supplementary Appendix~A.}
\label{fig:patch_extraction_examples}%
\vspace{-3mm}
\end{figure}
%%%%%% REPLACE BACK LATER %%%%%%
% Top: FAST keypoints (red) and selected keypoints (blue) from Algorithm~\ref{alg:patch_extraction} overlaid on IR (left) and RGB (right) images.

\begin{table*}[tb]
\caption{Patch matching FPR95 scores on the VIS-NIR benchmark
\cite{aguilera2016learning}. Scores are reported for each category and the
final mean. The best results are highlighted in \textbf{bold}.}
\label{tbl:NIR_results}%
\vspace{-2pt} \centering
\renewcommand{\arraystretch}{1.2}
\resizebox{0.95\textwidth}{!}{\fontsize{11}{13}\selectfont
		\begin{tabular}{l*{9}{c}}
			\toprule
			\textbf{Models} \thinspace & \thinspace \textbf{Field} \thinspace & \thinspace \textbf{Forest} \thinspace & \thinspace \textbf{Indoor} \thinspace & \thinspace \textbf{Mountain} \thinspace & \thinspace \textbf{Old Building} \thinspace & \thinspace \textbf{Street} \thinspace & \thinspace \textbf{Urban} \thinspace & \thinspace \textbf{Water} \thinspace & \thinspace \textbf{Mean} \\
			\midrule
			\multicolumn{10}{c}{\textbf{Traditional methods}}                                                                                                                       \\
			SIFT \cite{lowe2004distinctive} \thinspace & \thinspace 39.44 \thinspace & \thinspace 11.39 \thinspace & \thinspace 10.13 \thinspace & \thinspace 28.63 \thinspace & \thinspace 19.69 \thinspace & \thinspace 31.14 \thinspace & \thinspace 10.85 \thinspace & \thinspace 40.33 \thinspace & \thinspace 23.95 \\
			GISIFT \cite{firmenichy2011multispectral} \thinspace & \thinspace 34.75 \thinspace & \thinspace 16.63 \thinspace & \thinspace 10.63 \thinspace & \thinspace 19.52 \thinspace & \thinspace 12.54 \thinspace & \thinspace 21.80 \thinspace & \thinspace 7.21 \thinspace & \thinspace 25.78 \thinspace & \thinspace 18.60 \\
			LGHD \cite{aguilera2015lghd}   \thinspace & \thinspace 16.52 \thinspace & \thinspace 3.78 \thinspace & \thinspace 7.91 \thinspace & \thinspace 10.66 \thinspace & \thinspace 7.91 \thinspace & \thinspace 6.55 \thinspace & \thinspace 7.21 \thinspace & \thinspace 12.76 \thinspace & \thinspace 9.16 \\
			\midrule
			\multicolumn{10}{c}{\textbf{Descriptor learning}}                                                                                                                               \\
			Q-Net \cite{aguilera2017cross} \thinspace & \thinspace 17.01 \thinspace & \thinspace 2.70 \thinspace & \thinspace 6.16 \thinspace & \thinspace 9.61 \thinspace & \thinspace 4.61 \thinspace & \thinspace 3.99 \thinspace & \thinspace 2.83 \thinspace & \thinspace 8.44 \thinspace & \thinspace 6.86 \\
						L2-Net \cite{tian2017l2} \thinspace & \thinspace 16.67 \thinspace & \thinspace 0.76 \thinspace & \thinspace 2.07 \thinspace & \thinspace 5.98 \thinspace & \thinspace 1.89 \thinspace & \thinspace 2.83 \thinspace & \thinspace 0.62 \thinspace & \thinspace 11.11 \thinspace & \thinspace 5.25 \\
			Exp-TLoss \cite{wang2019better}  \thinspace & \thinspace 5.55 \thinspace & \thinspace 0.24 \thinspace & \thinspace 2.30 \thinspace & \thinspace 1.51 \thinspace & \thinspace 1.45 \thinspace & \thinspace 2.15 \thinspace & \thinspace 1.44 \thinspace & \thinspace 1.95 \thinspace & \thinspace 2.07 \\
			HardNet \cite{mishchuk2017working}      \thinspace & \thinspace 5.61 \thinspace & \thinspace 0.15 \thinspace & \thinspace 1.50 \thinspace & \thinspace 3.14 \thinspace & \thinspace 1.10 \thinspace & \thinspace 1.93 \thinspace & \thinspace 0.69 \thinspace & \thinspace 2.29 \thinspace & \thinspace 2.05 \\
			SOSNet \cite{tian2019sosnet} \thinspace & \thinspace 5.94 \thinspace & \thinspace 0.13 \thinspace & \thinspace 1.53 \thinspace & \thinspace 2.45 \thinspace & \thinspace 0.99 \thinspace & \thinspace 2.04 \thinspace & \thinspace 0.78 \thinspace & \thinspace 1.90 \thinspace & \thinspace 1.97 \\
			D-Hybrid-CL \cite{baruch2021joint} \thinspace & \thinspace 4.40 \thinspace & \thinspace 0.20 \thinspace & \thinspace 2.48 \thinspace & \thinspace 1.50 \thinspace & \thinspace 1.19 \thinspace & \thinspace 1.93 \thinspace & \thinspace 0.78 \thinspace & \thinspace 1.56 \thinspace & \thinspace 1.70 \\
		Transformer-Encoder \cite{moreshet2021paying}  \thinspace & \thinspace 4.22 \thinspace & \thinspace 0.13 \thinspace & \thinspace 1.48 \thinspace & \thinspace 1.03 \thinspace & \thinspace 1.06 \thinspace & \thinspace 1.03 \thinspace & \thinspace 0.90 \thinspace & \thinspace 1.90 \thinspace & \thinspace 1.44 \\
			HyNet \cite{tian2020hynet}  \thinspace & \thinspace 4.50 \thinspace & \thinspace \textbf{0.07} \thinspace & \thinspace 1.09 \thinspace & \thinspace 1.80 \thinspace & \thinspace 0.83 \thinspace & \thinspace \textbf{0.52} \thinspace & \thinspace 0.53 \thinspace & \thinspace 1.91 \thinspace & \thinspace 1.41 \\
			
			\textbf{Hyp-Net (Ours)} \thinspace & \thinspace \textbf{3.97} \thinspace & \thinspace 0.13 \thinspace & \thinspace 0.83 \thinspace & \thinspace 0.76 \thinspace & \thinspace 0.74 \thinspace & \thinspace 0.53 \thinspace & \thinspace \textbf{0.39} \thinspace & \thinspace 1.51 \thinspace & \thinspace \textbf{1.11} \\
% \textbf{Hyp-Net (SE)} \thinspace & \thinspace 4.77 \thinspace & \thinspace 0.13 \thinspace & \thinspace 0.96 \thinspace & \thinspace \textbf{0.65} \thinspace & \thinspace 0.70 \thinspace & \thinspace 0.55 \thinspace & \thinspace 0.43 \thinspace & \thinspace 1.53 \thinspace & \thinspace 1.21 \\
% \textbf{Hyp-Net (Full)} \thinspace & \thinspace 4.30 \thinspace & \thinspace 0.15 \thinspace & \thinspace \textbf{0.80} \thinspace & \thinspace 0.74 \thinspace & \thinspace \textbf{0.67} \thinspace & \thinspace 0.65 \thinspace & \thinspace 0.43 \thinspace & \thinspace \textbf{1.50} \thinspace & \thinspace 1.16 \\

                        \bottomrule
		\end{tabular}
	}
\vspace{-0mm}
\end{table*}

Only a handful of paired multimodal patch datasets exist, leading to saturated
results and limiting the exploration of key challenges such as domain transfer
and domain adaptation. To advance research in multimodal patch matching, we
introduce the \textbf{G}round \textbf{A}erial \textbf{P}atch \textbf{V}is-\textbf{IR}
(GAP-VIR) dataset, a new dataset of paired visible–infrared (VIS–IR) patches
derived from cross-platform imagery. The dataset contains over 500K multimodal
patches, both paired and unpaired, generated from 40K paired RGB and LWIR
(thermal) images sourced from six existing datasets
\cite{flir_dataset, liu2022target, jia2021llvip, han2023aerial, sun2022drone,
tang2022piafusion}, curated by Ran et al.~\cite{ran2025diffv2ir} as part of
the IR500K dataset. Full details of the patch-generation process are provided
in Appendix A.

%%%%%% REPLACE BACK LATER %%%%%%
%The full details on the patch dataset creation are available in Appendix~\ref{app:gap_vir_dataset}.

The images are divided into two categories: \textit{Aerial} and \textit{Ground-based} footage, yielding 270K and 235K patches, respectively. These categories differ not only in viewpoint, scene scale, and geometric distortions but also in resolution, level of detail, background clutter, and occlusion patterns. Moreover, camera and platform characteristics such as altitude, viewing angle, and motion further accentuate differences in both RGB and IR imagery. Examples are shown in Fig.~\ref{fig:patch_extraction_examples}, which presents paired RGB–IR images and their extracted patches from aerial and ground footage, highlighting variations in scale and appearance. By introducing two distinct subsets with substantial differences, the dataset supports both training and evaluation of adaptation methods under varied domain shifts.

\section{Experimental Results}
\label{sec:Experiments}

We evaluate our approach on multimodal patch-matching datasets. These datasets
consist of pairs of multimodal images and evaluation protocols that generate
matching and non-matching patches. We compare our method against prior
state-of-the-art approaches. Performance is measured using the false positive
rate at $95\%$ recall (FPR95), where lower values indicate better results.

\subsection{Datasets}

The VIS-NIR benchmark \cite{aguilera2016learning} contains 1.6 million pairs of visible and NIR $64\times64$ patches. It is divided into nine categories based on the VIS-NIR image sets from which the patches were derived. Following standard practice, only the first category, \textit{Country}, is used for training and evaluation with an 80\%–20\% split. The remaining eight categories (field, forest, indoor, mountain, old building, street, urban, and water) are reserved for testing.

The En et al.~\cite{en2018ts} benchmark consists of patches derived from three multimodal datasets. VeDAI~\cite{razakarivony2016vehicle}, a set of RGB and NIR images of environments and vehicles, was designed for vehicle detection in aerial imagery. The CUHK dataset \cite{wang2008face} contains facial photographs paired with matching sketches, and RGB-NIR~\cite{brown2011multi} consists of visible and NIR images. To extract patches and split the dataset into training, validation, and test sets, we follow the \cite{moreshet2021paying} protocol.
The VIS-LWIR benchmark, introduced by Aguilera et
al.~\cite{aguilera2016learning}, is a patch dataset consisting of 21,370
pair samples extracted from aligned visible and long-wave infrared (LWIR)
images. We follow the scheme of Yu et al.~\cite{yu2023feature} for selecting
non-matching patch pairs and for train-test separation, using an 80\%-20\% split. Unfortunately, we were unable to gain access to other datasets, such as the
one introduced by Zhang \cite{zhang2023learning}, because we did not receive a
response from the authors. In addition, we evaluated our models on the GAP-VIR dataset introduced in
Section~\ref{gap_vir_dataset}. We used the two patch categories, aerial and
ground-based, to evaluate cross-domain adaptation. For each model, we trained
on a single category, as well as on both categories
combined. We then evaluated each model on the test set of each category.

\begin{table}[tb]
\caption{Patch matching FPR95 scores on the three datasets of the En
et al. benchmark \cite{en2018ts}. The best results are highlighted in \textbf{bold}.}
\label{tbl:vedai_results}
\centering
\small
\begin{tabular}{lccc}
\toprule
\textbf{Model} \thinspace & \thinspace \textbf{VEDAI} \thinspace & \thinspace \textbf{CUHK} \thinspace & \thinspace \textbf{RGB-NIR} \\
\midrule
SIFT \cite{lowe2004distinctive} \thinspace & \thinspace 42.74 \thinspace & \thinspace 5.87 \thinspace & \thinspace 32.53 \\
MI-SIFT \cite{ma2010mi} \thinspace & \thinspace 11.33 \thinspace & \thinspace 7.34 \thinspace & \thinspace 27.71 \\
LGHD \cite{aguilera2015lghd} \thinspace & \thinspace 1.31 \thinspace & \thinspace 0.65 \thinspace & \thinspace 10.76 \\
\midrule
Siamese \cite{aguilera2016learning} \thinspace & \thinspace 0.84 \thinspace & \thinspace 3.38 \thinspace & \thinspace 13.17 \\
Pseudo-Siamese \cite{aguilera2016learning} \thinspace & \thinspace 1.37 \thinspace & \thinspace 3.7 \thinspace & \thinspace 15.6 \\
Q-Net \cite{aguilera2017cross} \thinspace & \thinspace 0.78 \thinspace & \thinspace 0.9 \thinspace & \thinspace 22.5 \\
TS-Net \cite{chen2009real} \thinspace & \thinspace 0.45 \thinspace & \thinspace 2.77 \thinspace & \thinspace 11.86 \\
Transformer \cite{moreshet2021paying} \thinspace & \thinspace \textbf{0} \thinspace & \thinspace 0.05 \thinspace & \thinspace 1.76 \\
\textbf{Hyp-Net (Ours)} \thinspace & \thinspace \textbf{0} \thinspace & \thinspace \textbf{0.03} \thinspace & \thinspace \textbf{1.39} \\
\bottomrule
\end{tabular}
\vspace{-0mm}
\end{table}

\begin{table}[tb]
\caption{Patch matching FPR95 scores on the VIS-LWIR benchmark
\cite{aguilera2016learning,yu2023feature} for descriptor-learning (left) and
similarity-learning (right) methods. The best results are highlighted in \textbf{bold}.
Our method matches the SOTA result reported by similarity-learning
methods, despite their limited practicality.}%
\label{tbl:lwir_results}
\centering
\begin{minipage}{0.48\linewidth}
\centering
\small
\begin{tabular}{lc}
\toprule
\multicolumn{2}{c}{\textbf{Descriptor learning}} \\
\midrule
\textbf{Model} & \textbf{FPR95} \\
\midrule
MatchNet \cite{han2015matchnet} & 9.83 \\
Siamese \cite{aguilera2016learning} & 13.89 \\
Pseudo-Siamese \cite{aguilera2016learning} & 10.34 \\
2-Channel \cite{zagoruyko2015learning} & 8.34 \\
\textbf{Hyp-Net (Ours)} & \textbf{0.51} \\
\bottomrule
\\
\\
\end{tabular}
\end{minipage}
\hfill\begin{minipage}{0.48\linewidth}
\centering
\small
\begin{tabular}{lc}
\toprule
\multicolumn{2}{c}{\textbf{Similarity learning}} \\
\midrule
\textbf{Model} & \textbf{FPR95} \\
\midrule
SCFDM \cite{quan2019cross} & 1.89 \\
MR-3A \cite{quan2021multi} & 1.56 \\
AFD-Net \cite{quan2019afd} & 1.43 \\
MFD-Net \cite{yu2022multibranch} & 0.97 \\
EFR-Net \cite{yu2023efficient} & 0.92 \\
FIL-Net \cite{yu2023feature} & 0.69 \\
KGL-Net \cite{yu2024and} & \textbf{0.51} \\
\bottomrule
\end{tabular}
\end{minipage}
\vspace{-3mm}
\end{table}

\subsection{Implementation Details}

\label{subsec:Implementation Details}

\begin{table*}[tb]
\caption{Patch matching FPR95 scores for cross-domain evaluation on the GAP-VIR dataset. Models are trained separately on the ground set (170K), the aerial set (180K), and the combined set, and evaluated on each test set individually. HardNet++ denotes HardNet~\cite{mishchuk2017working} with an enlarged architecture to match the number of convolutional channels in our model (see Table~\ref{tbl:network_architectures}). The best results are shown in \textbf{bold}.}
\label{tbl:GAP_VIR_results}
%\vspace{-2pt}
\centering
\renewcommand{\arraystretch}{1.2}
%\resizebox{0.95\textwidth}{!} {\fontsize{10}{10}\selectfont
\begin{tabular*}{\textwidth}{@{\extracolsep{\fill}} l *{7}{c}}
\toprule
% Row 1: Training groups only (no text in col 1 or Mean col)
\textbf{Model} & \multicolumn{2}{c}{\textbf{Training Set: Ground}} &
\multicolumn{2}{c}{\textbf{Training Set: Aerial}} &
\multicolumn{2}{c}{\textbf{Training Set: Combined}} &
\\
\cmidrule(lr){2-3}\cmidrule(lr){4-5}\cmidrule(lr){6-7}
% Row 2: diagbox + test-set columns + Mean
% \multicolumn{1}{c}{\diagbox[width=36mm,height=5mm]{\textbf{Model}}{\textbf{Test Set}}} &
% Ground & Aerial & Ground & Aerial & Ground & Aerial & Mean (combined only) \\
%\textbf{Model} \\
 \hfill\textbf{Test Set} & \textbf{Ground} & \textbf{Aerial} & \textbf{Ground} & \textbf{Aerial} & \textbf{Ground} & \textbf{Aerial} & \textbf{Mean (combined)} \\

\midrule
			HardNet++~\cite{mishchuk2017working}      \thinspace & \thinspace 3.94 \thinspace & \thinspace 6.10 \thinspace & \thinspace \textbf{12.28} \thinspace & \thinspace 2.04 \thinspace & \thinspace 4.39 \thinspace & \thinspace 2.09 \thinspace & \thinspace 3.24 \\
		Transformer~\cite{moreshet2021paying}  \thinspace & \thinspace 3.83 \thinspace & \thinspace 5.94 \thinspace & \thinspace 15.43 \thinspace & \thinspace 1.82 \thinspace & \thinspace 1.86 \thinspace & \thinspace 1.13 \thinspace & \thinspace 1.50 \\
			\textbf{Hyp-Net (Ours)} \thinspace & \thinspace 3.40 \thinspace & \thinspace \textbf{5.07} \thinspace & \thinspace 17.86 \thinspace & \thinspace 1.44 \thinspace & \thinspace 1.21 \thinspace & \thinspace 0.70 \thinspace & \thinspace 0.95 \\
			% \textbf{Hyp-Net (SE)} \thinspace & \thinspace 2.32 \thinspace & \thinspace 5.90 \thinspace & \thinspace 18.02 \thinspace & \thinspace \textbf{1.20} \thinspace & \thinspace 1.07 \thinspace & \thinspace 0.37 \thinspace & \thinspace 0.72 \\
			% \textbf{Hyp-Net (Full)} \thinspace & \thinspace \textbf{1.74} \thinspace & \thinspace 6.46 \thinspace & \thinspace 18.83 \thinspace & \thinspace 1.25 \thinspace & \thinspace \textbf{1.06} \thinspace & \thinspace \textbf{0.29} \thinspace & \thinspace \textbf{0.67} \\
			%             \bottomrule

        \end{tabular*}
\vspace{-0mm}
\end{table*}

We trained the Siamese network from scratch using the Adam optimizer without
weight decay, with parameters $\beta_{1}=0.9$ and $\beta_{2}=0.999$. We
employed a reduce-on-plateau learning rate schedule, reducing the learning rate by a
factor of $10$ when the evaluation loss did not improve for three consecutive
epochs. The learning rate started at $\eta_{max}=10^{-2}$ and decayed to
$\eta_{min}=10^{-5}$. In addition, we applied a linear warm-up schedule of
four epochs, starting at a learning rate of $0.25\times10^{-2}$ before
transitioning to the main schedule.

We repeated the learning rate schedule for four cycles, reinitializing the Adam optimizer at each cycle, following \cite{smith2017cyclical}, which showed the benefit of cyclical over monotonically decreasing schedules. The cycles correspond to the negative sample
mining strategies described in Section~\ref{ref:loss_function}, with random
sampling in the first cycle and hard negative mining in the remaining cycles.
For the VIS-LWIR benchmark, we employed 12 cycles due to the smaller number of
patches. This setup provides sufficient iterations for the model to adapt to
augmentations and dropout without modifying other training parameters.

The augmentations used during training include random horizontal flips, random
$90^{\circ}$ rotations, and random vertical flips, all applied identically to
both patches in a pair. In contrast, small random rotations of up to
$5^{\circ}$ and random gamma adjustments were applied independently to each
patch in a pair. The first group was applied identically to both patches
because the architecture, being relatively shallow, is sensitive to spatial
position. The effects of the first augmentation group are evaluated in the
ablation study (Table~\ref{tbl:ablation_results}).

\subsection{Results}

\label{subsec:results}

In Table~\ref{tbl:NIR_results}, we present our results on the VIS-NIR dataset,
measured using the FPR95 metric. As discussed in
Section~\ref{subsec: descriptor vs metric}, we restrict our evaluation to
descriptor-learning methods, comparing against both traditional approaches
\cite{lowe2004distinctive, firmenichy2011multispectral, aguilera2015lghd} and
recent deep learning methods \cite{aguilera2017cross, tian2017l2,
wang2019better, mishchuk2017working, tian2019sosnet, baruch2021joint,
moreshet2021paying, tian2020hynet}. We demonstrate that our architecture significantly outperforms previous works by the overall mean score and across most categories. We achieves a mean
score of $1.11$, improving upon the previous best of $1.41$ and achieving up
to a 30\% reduction in FPR95 across some categories. As test performance
fluctuates on unseen data, we report the score at the end of training.

The results on the En et al.~\cite{en2018ts} benchmark are shown in Table~\ref{tbl:vedai_results}. We achieve a new improved state of the art (SOTA) FPR95 scores on the CUHK and NIR-RGB datasets, and match the zero FPR95 on VeDAI, previously reported in~\cite{moreshet2021paying}. This outcome stems from the training and test sets in this dataset sharing similar distributions.

Table~\ref{tbl:lwir_results} presents results for the VIS-LWIR benchmark. In this benchmark, we show a substantial improvement over all descriptor-learning methods, reducing the FPR95 from $8.34$ reported by Han et al.~\cite{han2015matchnet} to $0.51$ (a $94\%$ reduction). For reference, we also report recent scores for metric learning methods. Although metric learning methods enjoy a built-in advantage at inference, making them impractical at scale, our method achieves comparable performance to the recently reported SOTA by Yu et al.~\cite{yu2024and}.

Table~\ref{tbl:GAP_VIR_results} presents GAP-VIR results from training on
single or combined domains, with the former enabling cross-domain evaluation.
We observe that Hyp-Net consistently outperforms the other models, achieving
FPR95 improvements of 10\%–60\% across all domains except one. When examining
cross-domain generalization, Hyp-Net trained on ground-based imagery
outperforms other models on its own test set and when transferred to the
\emph{aerial} set. However, when trained on {aerial} imagery, both Hyp-Net and
competing models struggle to generalize to {ground} data. This is likely due to
substantial domain gap and the lower diversity of aerial patches, as
illustrated in Fig.~\ref{fig:patch_extraction_examples}. Finally, when trained
on both domains, attention-based models show significant improvements,
suggesting that increased data diversity enhances generalization.

\begin{table*}[tb]
\caption{Ablation study on backbone architectures for patch matching. \textit{ViT Small} and \textit{Swin Small} denote reduced versions with fewer layers, while \textit{ResNet-18 L3} and \textit{ResNet-18 L2} refer to ResNet-18 with only three or two residual layers, respectively. Architectural details are provided in Table~2 of Appendix~B in the supplementary material. Results are reported as FPR95 scores on the VIS-NIR dataset, both per category and as overall means. Best results are shown in \textbf{bold}, and second-best in \underline{underline}.}%
\label{tbl:architecture_results}
%\vspace{-2pt}
\centering
\renewcommand{\arraystretch}{1.2}
\resizebox{0.95\textwidth}{!}{\fontsize{11}{13}\selectfont
		\begin{tabular}{l*{9}{c}}
			\toprule
			\textbf{Model} \thinspace & \thinspace \textbf{Field} \thinspace & \thinspace \textbf{Forest} \thinspace & \thinspace \textbf{Indoor} \thinspace & \thinspace \textbf{Mountain} \thinspace & \thinspace \textbf{Old Building} \thinspace & \thinspace \textbf{Street} \thinspace & \thinspace \textbf{Urban} \thinspace & \thinspace \textbf{Water} \thinspace & \thinspace \textbf{Mean} \\
			\midrule
			VIT~\cite{dosovitskiy2020image} \thinspace & \thinspace 10.24 \thinspace & \thinspace 0.43 \thinspace & \thinspace 2.57 \thinspace & \thinspace 3.48 \thinspace & \thinspace 1.03 \thinspace & \thinspace 1.52 \thinspace & \thinspace 0.92 \thinspace & \thinspace 3.29 \thinspace & \thinspace 2.93 \\
			VIT Small \thinspace & \thinspace 8.80 \thinspace & \thinspace 0.23 \thinspace & \thinspace 2.21 \thinspace & \thinspace 2.64 \thinspace & \thinspace 1.42 \thinspace & \thinspace 1.76 \thinspace & \thinspace 0.81 \thinspace & \thinspace 2.19 \thinspace & \thinspace 2.51 \\
\midrule
SWIN~\cite{liu2021swin} \thinspace & \thinspace 7.59 \thinspace & \thinspace 0.95 \thinspace & \thinspace 1.94 \thinspace & \thinspace 2.25 \thinspace & \thinspace 1.29 \thinspace & \thinspace 2.53 \thinspace & \thinspace 0.78 \thinspace & \thinspace 2.23 \thinspace & \thinspace 2.44 \\
			SWIN Small \thinspace & \thinspace \underline{4.48} \thinspace & \thinspace 0.12 \thinspace & \thinspace 1.08 \thinspace & \thinspace 1.05 \thinspace & \thinspace 0.91 \thinspace & \thinspace 1.14 \thinspace & \thinspace 0.63 \thinspace & \thinspace 1.63 \thinspace & \thinspace 1.38 \\
			    		\midrule
ResNet-18 \thinspace & \thinspace  13.23 \thinspace & \thinspace 8.42 \thinspace & \thinspace 4.22 \thinspace & \thinspace 7.73 \thinspace & \thinspace 3.93 \thinspace & \thinspace 6.24 \thinspace & \thinspace 2.68 \thinspace & \thinspace 5.21 \thinspace & \thinspace 6.46 \\
ResNet-18 L3 \thinspace & \thinspace  4.67 \thinspace & \thinspace 0.46 \thinspace & \thinspace 1.77 \thinspace & \thinspace 1.53 \thinspace & \thinspace 0.96 \thinspace & \thinspace 1.34 \thinspace & \thinspace 0.82 \thinspace & \thinspace 1.96 \thinspace & \thinspace 1.69 \\
ResNet-18 L2  \thinspace & \thinspace  5.63 \thinspace & \thinspace \underline{0.11} \thinspace & \thinspace \underline{1.04} \thinspace & \thinspace 0.92 \thinspace & \thinspace 0.88 \thinspace & \thinspace 0.80 \thinspace & \thinspace 0.79 \thinspace & \thinspace 1.72 \thinspace & \thinspace 1.48 \\
\midrule
Hyp-Net Small \thinspace & \thinspace  5.93 \thinspace & \thinspace \textbf{0.08} \thinspace & \thinspace 1.10 \thinspace & \thinspace \textbf{0.72} \thinspace & \thinspace \underline{0.82} \thinspace & \thinspace \textbf{0.48} \thinspace & \thinspace 0.63 \thinspace & \thinspace \textbf{1.46} \thinspace & \thinspace 1.40 \\
			Hyp-Net Large \thinspace & \thinspace 4.54 \thinspace & \thinspace 0.37 \thinspace & \thinspace 1.21 \thinspace & \thinspace 0.91 \thinspace & \thinspace \underline{0.82} \thinspace & \thinspace 0.86 \thinspace & \thinspace \underline{0.48} \thinspace & \thinspace 1.54 \thinspace & \thinspace \underline{1.34} \\
			\textbf{Hyp-Net (Ours)} \thinspace & \thinspace \textbf{3.97} \thinspace & \thinspace 0.13 \thinspace & \thinspace \textbf{0.83} \thinspace & \thinspace \underline{0.76} \thinspace & \thinspace \textbf{0.74} \thinspace & \thinspace \underline{0.53} \thinspace & \thinspace \textbf{0.39} \thinspace & \thinspace \underline{1.51} \thinspace & \thinspace \textbf{1.11} \\

                        \bottomrule
		\end{tabular}
	}
\vspace{-3mm}
\end{table*}

\subsection{Ablation studies}

\label{subsec:ablation_studies}

\begin{table}[tbh]
\caption{Ablation study of architectural elements: Patch matching FPR95 scores
on the VIS-NIR dataset, where we vary a single architectual element at a time: augmentations, normalization blocks, or hypernetwork modules. The best result is highlighted in \textbf{bold}.}%
\label{tbl:ablation_results}
\centering
\small
\begin{tabular}{lcccc}
\toprule
% \shortstack{Ablation\\ Check} & \shortstack{Hypernet\\ Model} & \shortstack{Norm \\ \vspace{7pt}} & \shortstack{$90^\circ$ Rotation\\ \& Vertical Flip} & \shortstack{FPR95 \\ \vspace{7pt}}  \\
\textbf{\shortstack{Ablation\\ Check}} &
\textbf{\shortstack{Hypernet\\ Model}} &
\textbf{\shortstack{Norm\\\vphantom{aaa}}} &
\shortstack{$\bm{90^\circ}$ \\ \textbf{Rotation} \\ \textbf{Vertical Flip}} &
\textbf{\shortstack{FPR95\\\vphantom{aaa}}} \\
\midrule
Augmentations  & +        & CIN  & --       & 1.21 \\
Normalization  & +        & BN   & +        & 1.26 \\
Normalization  & +        & IN   & +        & 1.19 \\
Architecture   & --       & CIN  & +        & 1.30 \\
\textbf{Hyp-Net} & +      & CIN  & +        & \textbf{1.11} \\
\bottomrule
\end{tabular}
\vspace{-0mm}
\end{table}

%%%%%% REPLACE BACK LATER %%%%%%
% Architectural details are provided in Table~\ref{tbl:ablation_architectures}.

\noindent\textbf{Architectural Components.} We conducted an ablation study on
the architectural elements of Hyp-Net, and the results are shown in
Table~\ref{tbl:ablation_results}. We observe the best performance when all
elements are used together, suggesting that they play complementary roles in
the matching process. The addition of hypernetwork modules improves the FPR95
score by $17\%$, demonstrating their effectiveness in capturing relevant
higher-order information. The architecture with CIN blocks outperforms those
with IN and BN blocks by $7\%$ and $14\%$, respectively. This provides
evidence that IN blocks are more robust to appearance changes (as suggested in~\cite{quan2019afd}), and that the modality-based information
captured by CIN blocks is additionally beneficial. Random $90^{\circ}$
rotation and vertical flip augmentations also improve the score by $9\%$,
suggesting that encouraging rotation-invariant features enhances robustness on
unseen data.

\noindent\textbf{Backbone.} We conducted ablation studies to assess the
effectiveness of different backbones on the VIS-NIR dataset, including
ResNet~\cite{he2016deep}, ViT~\cite{dosovitskiy2020image}, and
Swin~\cite{liu2021swin} Transformers. We also evaluated the effect of backbone
size (layers and parameters), with full details provided in Appendix B.
Results are shown in Table~\ref{tbl:architecture_results}. First, we find that
neither larger nor smaller convolutional backbones improve generalization,
yielding FPR95 scores of $1.34$ and $1.40$, compared to Hyp-Net's $1.11$. For
ResNet-18, the full model performs poorly, but reducing depth improves
results, with the two-layer reduction achieving an FPR score of $1.48$.

%%%%%% REPLACE BACK LATER %%%%%%
% We also evaluated the effect of backbone size (layers and parameters), with full details provided in Appendix~\ref{sec:alternative_backbones}.

ViT variants perform poorly overall, suggesting that despite their strong
performance in other vision tasks, they may be less effective for
generalization in patch matching. Swin performs better than ViT, with Swin
Small reaching $1.38$, slightly surpassing the previous SOTA of
$1.41$, though still behind Hyp-Net. Overall, these experiments indicate that
(a) smaller networks up to a certain size generalize better, and (b) newer
architectures do not necessarily improve performance on the diverse VIS-NIR
test set.

\section{Conclusion}

We introduced Hyp-Net, a hypernetwork-based CNN architecture for
multimodal patch matching, and GAP-VIR, a new RGB-IR patch dataset spanning
both ground and aerial imagery. Our approach achieves state-of-the-art results
on VIS-NIR, VIS-LWIR, and En et al. benchmark datasets, using an architecture
comparable in size to previous solutions. Through ablation studies, we
analyzed the role of backbone choice and observed that smaller CNNs tend to
generalize better than standard-size CNNs and transformer-based models.
GAP-VIR further enables evaluation under domain adaptation scenarios,
providing a challenging benchmark for future research. We hope this work
stimulates further research on cross-modal patch correspondence and broader
multimodal matching tasks.

{
    \small
    \bibliographystyle{ieeenat_fullname}
    \bibliography{main}
}


\clearpage
\appendix
\section*{Supplementary Material} % Title for the supplement
% \documentclass[10pt,twocolumn,letterpaper]{article}
% \usepackage[review]{cvpr} 
% \input{preamble}
% \usepackage[pagebackref,breaklinks,colorlinks]{hyperref}
% \def\paperID{} % *** Enter the Paper ID here
% \def\confName{CVPR}
% \def\confYear{2026}

% \title{Multi-Sensor Matching with HyperNetworks}

% \begin{document}
% \maketitlesupplementary
% \appendix

\section{GAP-VIR Dataset Generation}

\label{app:gap_vir_dataset}

%%%%%% REPLACE BACK LATER %%%%%%
% As introduced in Sec.~\ref{gap_vir_dataset}
As introduced in Sec.~IV, GAP-VIR is a dataset of paired
VIS-IR patches derived from cross-platform imagery. Here, we provide additional
details on its construction. The patches were created from 40K paired VIS-IR
images of the IR500K dataset introduced by Ran et al.~\cite{ran2025diffv2ir}.
IR500K is composed of multiple infrared and visible image datasets, from which
we used a subset of paired images captured by ground-based and aerial RGB and
LWIR (thermal) cameras. Table~\ref{tbl:paired_datasets} summarizes the six
original paired multimodal datasets \cite{flir_dataset, liu2022target,
jia2021llvip, han2023aerial, sun2022drone, tang2022piafusion}.
Table~\ref{tbl:paired_datasets} summarizes the paired multimodal datasets used
to construct our patch dataset, including imaging platform, acquisition type,
and splitting protocols. We also indicate whether grouping information was
available to guide data diversity and whether cross-frame duplicate removal
was applied.

Our split protocol is designed to minimize the risk of similarity leakage
between train and test sets by taking into account acquisition type, grouping,
and sequential order. All datasets were divided using a 70\%/10\%/20\%
train/validation/test ratio, with grouping applied first where relevant. For
LLVIP~\cite{jia2021llvip}, whose data are organized into sets of frames from
individual static cameras, we grouped by camera so that all frames from the
same camera remained in a single split. For MSRS~\cite{tang2022piafusion},
image filenames include a day/night indicator; we first separated the dataset
into daily and nightly subsets and then split each subset sequentially into
train, validation, and test. This choice was motivated by
M3FD~\cite{liu2022target}, which also contains daily and nightly imagery but
lacks explicit indicators for separation. After applying these grouping rules,
all splits were finalized using sequential (lexical) file order, ensuring
consistent treatment across datasets. These steps together help prevent
visually similar or identical patches from appearing in both training and test sets.

\begin{table*}[tb]
\caption{Summary of paired multimodal datasets used to construct our
multimodal patch dataset. For each dataset, we report the imaging platform
(ground or aerial), the number of paired samples, acquisition type,
availability of grouping information (e.g., by camera or time of day) for
diversity in splits, the protocol for train/validation/test division, and
whether cross-frame duplicate patch removal was applied.}%
\label{tbl:paired_datasets}
\begin{center}
{\small 
\begin{tabular}{lccccccc}
\toprule
\shortstack{Dataset\\ Name} & \shortstack{Imaging \\ Platform} & \shortstack{Number of \\ Image Pairs}  & \shortstack{Acquisition \\ Type} & \shortstack{Existing \\ Grouping} & \shortstack{Data Split \\ Protocol} & \shortstack{Patch  \\ Deduplication}  \\
\midrule
AVIID~\cite{han2023aerial} \thinspace & \thinspace Aerial \thinspace & \thinspace 1{,}280 \thinspace & \thinspace Continuous Video \thinspace & \thinspace - \thinspace & \thinspace Sequential \thinspace & \thinspace - \\
Visdrone~\cite{sun2022drone} \thinspace & \thinspace Aerial \thinspace & \thinspace 13{,}442 \thinspace & \thinspace Multi-Scene Frames \thinspace & \thinspace Train/Val/Test \thinspace & \thinspace Already Split \thinspace & \thinspace - \\
LLVIP~\cite{jia2021llvip} \thinspace & \thinspace Ground \thinspace & \thinspace 15{,}488 \thinspace & \thinspace Fixed-Camera Frame Sets \thinspace & \thinspace Camera Sets \thinspace & \thinspace Camera sets \thinspace & \thinspace + \\
FLIR~\cite{flir_dataset} \thinspace & \thinspace Ground \thinspace & \thinspace 5{,}142 \thinspace & \thinspace Sampled Video Frames \thinspace & \thinspace - \thinspace & \thinspace Sequential \thinspace & \thinspace - \\
M3FD~\cite{liu2022target} \thinspace & \thinspace Ground \thinspace & \thinspace 4{,}200 \thinspace & \thinspace Multi Scene Frames \thinspace & \thinspace - \thinspace & \thinspace Sequential \thinspace & \thinspace - \\
MSRS~\cite{tang2022piafusion} \thinspace & \thinspace Ground \thinspace & \thinspace 1{,}444 \thinspace & \thinspace Multi Scene Frames \thinspace & \thinspace Day/Night \thinspace & \thinspace Day/Night Sequential \thinspace & \thinspace - \\
\bottomrule
\end{tabular}
}
\end{center}
\par
%\par
%\vskip -0.1in
\end{table*}

After defining the dataset splits, we turn to the process of patch extraction.
Several factors must be considered when extracting multimodal patches:
spectral differences, detector robustness, feature bias (e.g., foliage), and
duplicate removal. Details on these aspects are scarce in the literature, with
prior work typically providing only high-level descriptions
\cite{aguilera2016learning}. We therefore introduce a detailed patch-level
extraction procedure tailored for multimodal datasets. We detect keypoints
independently per modality using FAST~\cite{rosten2008faster} and rank them by
the Harris corner score~\cite{harris1988combined}. To reduce spatial
redundancy, we group keypoints into $16\times16$ cells and retain only those
that are present (or have neighbors) in both modalities. For $64\times64$
patches, we apply non-maximum suppression (NMS) both within each modality and
across modalities. For LLVIP static-camera sets, we further de-duplicate
patches across frames by discarding those whose ORB
descriptors~\cite{rublee2011orb} match those already included. The complete
cross-spectral patch extraction procedure is given in
Algorithm~\ref{alg:patch_extraction}.

%\newcommand{\Nbr}[1]{\ensuremath{\mathcal{N}_{3\times3}\!\left(#1\right)}}
%\noindent\textbf{Notation.} For a grid cell $c$, $\Nbr(c)$ denotes the 3$\times$3 Moore neighborhood (cell $c$ and its 8 adjacent cells).

\begin{algorithm}[tb]
\caption{Cross-spectral patch extraction per image pair}
\label{alg:patch_extraction}
\begin{algorithmic}[1]
\State \textbf{Input:} Paired images $I^{(a)}, I^{(b)}$; patch size $P{=}64$; grid cell size $g{=}16$; margin $m{=}P/2$; Harris threshold factor $\tau_H$; intra-/inter-modality IoU thresholds $t_\mathrm{intra}, t_\mathrm{inter}$; target patches per pair $N$
\State \textbf{Output:} Selected patch set $\mathcal{S}$
\For{$u \in \{a,b\}$} \Comment{Per modality}
\State $R \gets \textsc{Harris}(I^{(u)})$; $K^{(u)} \gets \textsc{FastDetect}(I^{(u)})$; set $k.\mathrm{score} \leftarrow R(k)$ for all $k \in K^{(u)}$
\State \textbf{Remove edge points:} delete $k \in K^{(u)}$ with border distance $< m$
\State \textbf{Remove low-score points:} let $s_{\max} \gets \max_{k \in K^{(u)}} k.\mathrm{score}$; delete any $k$ with $k.\mathrm{score} < \tau_H s_{\max}$
\State \textbf{Partition and keep top per cell:} divide image into $g{\times}g$ cells; in each cell, retain only the highest-score keypoint
\State $C^{(u)} \gets \{\textsc{Cell}(k) \mid k \in K^{(u)}\}$ \Comment{Cells occupied after filtering}
\EndFor
\State $C \gets \mathcal{N}_{3\times3}(C^{(a)}) \cap \mathcal{N}_{3\times3}(C^{(b)})$ \Comment{Expand each cell to its 8 neighbors (Moore neighborhood) and intersect modalities}
\For{$u \in \{a,b\}$}
\State \textbf{Cross-modal consistency:} keep only $k \in K^{(u)}$ with $\textsc{Cell}(k) \in C$
\State Extract a $P{\times}P$ patch around each $k \in K^{(u)}$ to form $\mathcal{P}^{(u)}$
\State \textbf{Intra-modal NMS:} scan $\mathcal{P}^{(u)}$ by descending score; retain a patch if its IoU with all kept patches is $< t_\mathrm{intra}$
\State Sort $\mathcal{P}^{(u)}$ by score; select $\lfloor N/4 \rfloor$ top patches and sample $\lceil N/4 \rceil$ additional ones without replacement from the remainder
\EndFor
\State \textbf{Optional de-duplication (static frames):} per modality, compare each patch’s ORB descriptor against patches from previous frames in the same or neighboring grid cell; discard the patch if a close match is found, otherwise keep it in the cache
\State $\mathcal{S} \gets \mathcal{P}^{(a)} \cup \mathcal{P}^{(b)}$
\State \textbf{Inter-modal NMS:} remove any patch in $\mathcal{S}$ whose IoU with a higher-scoring kept patch is $\ge t_\mathrm{inter}$
\State \Return $\mathcal{S}$
\end{algorithmic}
\end{algorithm}

\begin{table*}[tb]
\caption{Backbone architectures used in the ablation study. We compare
structural details of ResNet-18, ViT, Swin Transformer, and our custom CNN
backbones (standard, small, and large variants). \textit{ViT Small} and
\textit{Swin Small} are reduced versions with fewer layers, while
\textit{ResNet-18 L3} and \textit{ResNet-18 L2} use only three or two residual
layers, respectively. The \textit{Embedding Channels} column reports the
embedding size for Transformers and the output channels of the final CNN
layer.}%
\label{tbl:ablation_architectures}
\begin{center}
{\small \textsc{
\begin{tabular}{lcccccccc}
\toprule
\shortstack{Model \\ \vspace{5pt}} &
\shortstack{Hypernet \\ Module} &
\shortstack{Norm \\ \vspace{5pt}} &
\shortstack{Layers \\ \vspace{5pt}} &
\shortstack{Embedding \\ Channels} &
\shortstack{Num \\ Heads} &
\shortstack{Window \\ Size} &
\shortstack{Patch \\ Size} &
\shortstack{Num \\ Parameters} \\
\midrule
ViT~\cite{dosovitskiy2020image} & - & LN & 12 & 768 & 12 & - & 16 & 51{,}942{,}784 \\
ViT Small & - & LN & 6 & 768 & 12 & - & 16 & 26{,}730{,}880 \\
\midrule
Swin~\cite{liu2021swin} & - & LN & [2,2,6,2] & 384 & [3,6,12,24] & [4,4,4,2] & 4 & 27{,}893{,}994 \\
Swin Small & - & LN & [2,1,1] & 384 & [3,6,12] & [8,4,4] & 4 & 3{,}614{,}296 \\
\midrule
ResNet-18 & - & BN & 17 & 512 & - & - & - & 11{,}442{,}388 \\
ResNet-18 L3 & - & BN & 13 & 256 & - & - & - & 3{,}305{,}679 \\
ResNet-18 L2 & - & BN & 9 & 128 & - & - & - & 2{,}240{,}690 \\
\midrule
Hyp-Net Small & + & CIN & 6 & 128 & - & - & - & 1{,}350{,}435 \\
Hyp-Net Large & + & CIN & 10 & 256 & - & - & - & 4{,}111{,}206 \\
Hyp-Net (Ours) & + & CIN & 8 & 128 & - & - & - & 1{,}662{,}820 \\
\bottomrule
\end{tabular}
}}
\end{center}
\par
\end{table*}

\section{Alternative Backbones}

\label{sec:alternative_backbones}

%%%%%% REPLACE BACK LATER %%%%%%
% As discussed in Sec.~\ref{subsec:ablation_studies}
% All variants were trained using the same protocol described in Subsection~\ref{subsec:Implementation Details}

As discussed in Sec.~V-D, our ablation studies
compared CNN and Transformer architectures using both standard and custom
backbones. Table~\ref{tbl:ablation_architectures} summarizes all tested
backbone configurations. For the standard CNN backbone underlying Hyp-Net,
similar in structure and size to those used in prior work \cite{tian2020hynet,
moreshet2021paying, baruch2021joint, tian2019sosnet, mishchuk2017working}, we
created larger and smaller variants by adding or removing two layers,
respectively. For a ResNet backbone, we used the compact ResNet-18, which
consists of four residual blocks with two convolutions each. We created three
variants with four, three, and two residual blocks, the smallest having a
parameter count comparable to the standard CNN backbone. All variants were
trained using the same protocol described in Sec.~V-B.

When using default Transformer architectures with a nonstandard patch size
($64\times64$ instead of the usual $224\times224$), some adjustments were
unavoidable. We adapted the positional encodings accordingly and modified the
local window size for the Swin architecture. Given that our Hyp-Net CNN has far
fewer parameters than ViT and Swin, we also introduced smaller variants
(\textit{ViT Small} and \textit{Swin Small}). We did not extend the
Transformer architectures with hypernetworks, as they already include built-in
attention mechanisms. For training the ViT and Swin architectures, we largely
followed the procedures in Section~5.2,
with minor modifications. In particular, we used a cosine annealing
scheduler~\cite{loshchilov2016sgdr} with a 50-epoch cycle for the smaller
variants but only 2 epochs for the standard Transformer architectures.

%%% RESOTRE LATER  Subsection~\ref{subsec:Implementation Details}

% {
%     \small
%     \bibliographystyle{ieeenat_fullname}
%     \bibliography{main}
% }

% \end{document}
 % This pulls in the content of supp.tex

\end{document}